\documentclass{article}

\newif\ifpreprintversion
\preprintversiontrue

\newif\ifshowcorrespondence
\showcorrespondencefalse

\usepackage{microtype}
\usepackage{graphicx}
\usepackage{subcaption}

\usepackage{booktabs} 
\usepackage{multirow} 

\usepackage{hyperref}
\usepackage{comment} 

\ifpreprintversion
  \usepackage[preprint]{icml2026}
\else
  \usepackage{icml2026}
\fi

\usepackage{amsmath}
\usepackage{amssymb}
\usepackage{mathtools}
\usepackage{amsthm}

\usepackage[capitalize,noabbrev]{cleveref}

\theoremstyle{plain}

\theoremstyle{definition}

\theoremstyle{remark}

\newtheorem{observation}{Observation}[section]

\usepackage[disable,textsize=tiny]{todonotes}
\usepackage{colortbl}
\definecolor{tancolor}{RGB}{187,130,90}
\definecolor{slateblue}{RGB}{85,104,154}
\definecolor{lm_purple}{RGB}{227,227,240}
\definecolor{lm_purple_low}{RGB}{240,240,248}

\icmltitlerunning{TriAttention: Efficient Long Reasoning with Trigonometric KV Compression}

\begin{document}

\twocolumn[
  \icmltitle{TriAttention: Efficient Long Reasoning with Trigonometric KV Compression}

  \icmlsetsymbol{equal}{*}

  \ifpreprintversion
    \begin{icmlauthorlist}
      \icmlauthor{\quad\quad Weian Mao$^{\,1*}$ \quad\quad}{}
      \icmlauthor{Xi Lin$^{\,3*}$ \quad\quad}{}
      \icmlauthor{Wei Huang$^{\,2*}$ \quad\quad}{}
      \icmlauthor{Yuxin Xie}{footnote}
      \icmlauthor{\quad\quad Tianfu Fu$^{\,1}$ \quad\quad}{}
      \icmlauthor{\quad\quad Bohan Zhuang$^{\,3}$ \quad\quad}{}
      \icmlauthor{Song Han$^{\,1\,2}$ \quad\quad}{}
      \icmlauthor{Yukang Chen$^{\,2}$}{}
    \end{icmlauthorlist}

    \icmlaffiliation{footnote}{MIT \quad $^2$NVIDIA \quad $^3$ZJU \quad $^*$Equal contribution}

    \icmlcorrespondingauthor{Yukang Chen}{yukangc@nvidia.com}
  \else
    \begin{icmlauthorlist}
      \icmlauthor{Anonymous Author(s)}{anon}
    \end{icmlauthorlist}

    \icmlaffiliation{anon}{Anonymous Institution}

    \icmlcorrespondingauthor{Anonymous Author}{anon.email@domain.com}
  \fi

  \icmlkeywords{KV Cache, LLM, Attention, RoPE, Compression, Reasoning}

  \vskip 0.3in
]

\printAffiliationsAndNotice{}

\begin{abstract}
Extended reasoning in large language models (LLMs) 
creates severe KV cache memory bottlenecks. Leading KV cache compression methods estimate KV importance using attention scores from recent post-RoPE queries. However, queries rotate with position during RoPE, making representative queries very few, leading to poor top-key selection and unstable reasoning.
To avoid this issue, we turn to the pre-RoPE space, where we observe that Q and K vectors are highly concentrated around fixed non-zero centers and remain stable across positions---\textit{Q/K concentration}. We show that this concentration causes queries to preferentially attend to keys at specific distances (e.g., nearest keys), with the centers determining which distances are preferred via a trigonometric series. Based on this, we propose TriAttention to estimate key importance by leveraging these centers. Via the trigonometric series, we use the distance preference characterized by these centers to score keys according to their positions, and also leverage Q/K norms as an additional signal for importance estimation. On AIME25 with 32K-token generation, TriAttention matches Full Attention reasoning accuracy while achieving 2.5$\times$ higher throughput or 10.7$\times$ KV memory reduction, whereas leading baselines achieve only about half the accuracy at the same efficiency.
\ifpreprintversion
TriAttention enables OpenClaw deployment on a single consumer GPU, where long context would otherwise cause out-of-memory with Full Attention. Our code is available at \url{https://github.com/WeianMao/triattention}.
\fi
\end{abstract}

\section{Introduction}

\begin{figure}[t]
  \vskip 0.1in
  \begin{center}
    \includegraphics[width=\columnwidth]{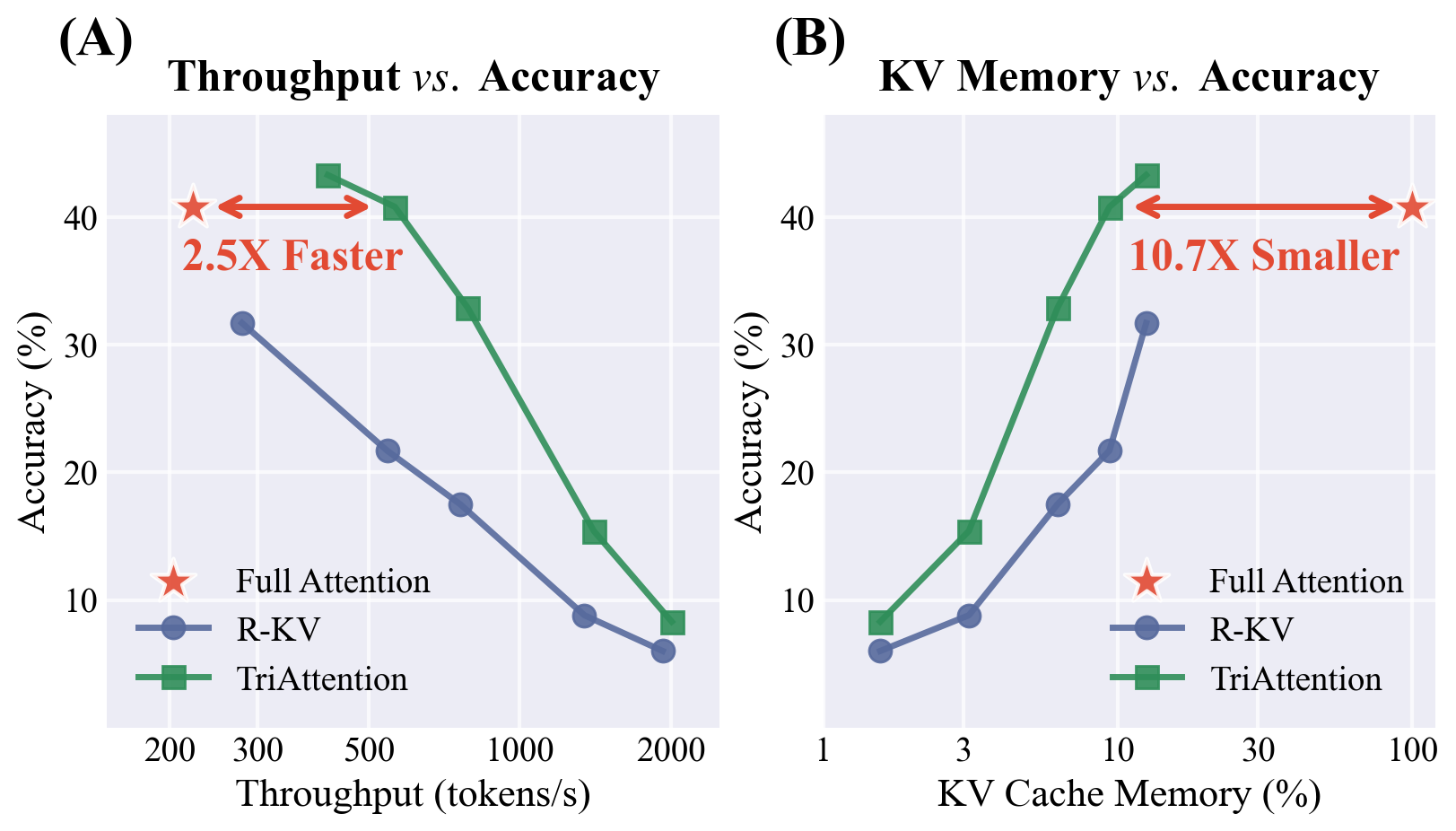}
    \caption{Performance trade-offs on AIME25 (Qwen3-8B). \textbf{(A)} At equivalent accuracy (40.8\%), TriAttention achieves 2.5$\times$ higher throughput than Full Attention. \textbf{(B)} TriAttention reduces KV cache memory by 10.7$\times$ while matching Full Attention accuracy.}
    \label{fig:teaser}
  \end{center}
  \vskip -0.2in
\end{figure}

Long reasoning in LLMs
produces chain-of-thought sequences spanning tens of thousands of tokens~\cite{wei2022cot,deepseek2025r1,shao2024deepseekmath,chen2025scaling}. KV cache grows proportionally, creating severe memory bottlenecks. KV cache compression addresses this by retaining only the most important tokens, with importance estimated from attention scores computed over recent queries~\cite{li2024snapkv,zhang2023h2o,devoto2025expectedattention,zhou2024survey,tang2024quest,shi2024keep}.

\begin{figure*}[t]
  \vskip 0.2in
  \begin{center}
    \includegraphics[width=0.95\textwidth]{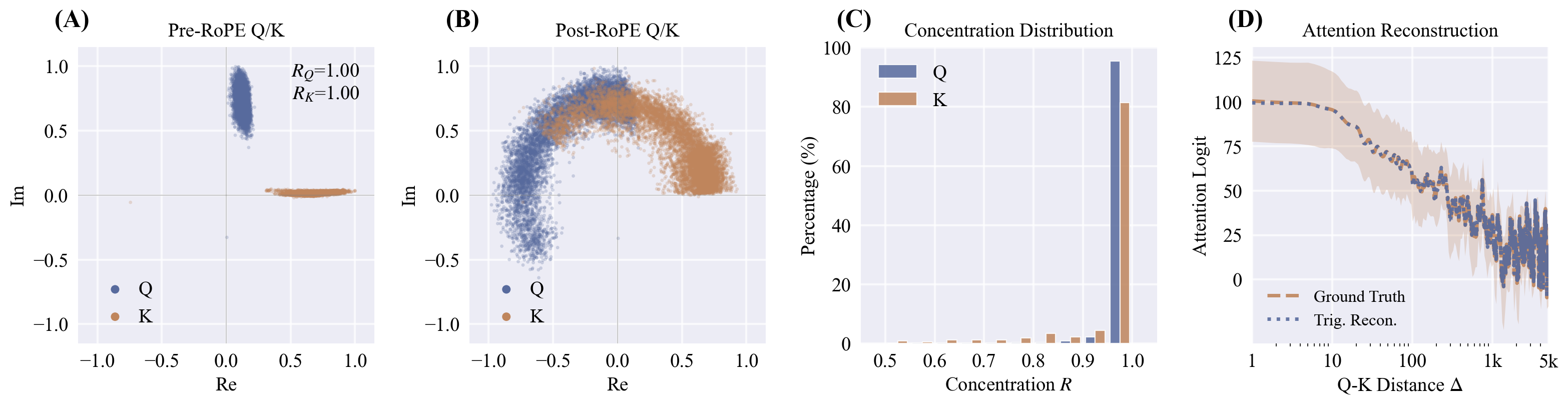}
    \caption{Q/K concentration and its implications for attention. \textbf{(A)} Pre-RoPE Q/K vectors at the dominant frequency band are highly concentrated (high Mean Resultant Length $R$). \textbf{(B)} RoPE rotation disperses these vectors into arc patterns. In (A-B), three distinct input sequences are overlayed, showing this structure is stable across content. \textbf{(C)} This concentration holds across nearly all heads. \textbf{(D)} When Q/K are concentrated, attention logits can be accurately reconstructed using a trigonometric series (Pearson $r = 0.72$).}
    \label{fig:intro-combined}
  \end{center}
  \vskip -0.2in
\end{figure*}

However, these methods are inherently unstable: only a few queries are usable for importance estimation. They operate on post-RoPE queries, which rotate with position as illustrated in Figure~\ref{fig:intro-combined}(B); consequently, only the most recent queries retain up-to-date orientations, forming a tiny observation window. With so few representative queries, important keys go undetected---a token receiving low attention during this short window may be permanently evicted, even if it becomes critical later. This is particularly challenging for retrieval heads~\cite{wu2025retrieval,xiao2025duoattention}, where relevant tokens can remain dormant for long periods before becoming essential. In reasoning, such loss breaks the chain of thought. Prior work confirms this limitation: increasing the observation window does not help---performance peaks at around 25 queries, a tiny fraction of typical long contexts, and declines thereafter~\cite{zhang2025lazyevictionlaggedkveviction}.

We therefore turn to the pre-RoPE space, where we observe a notable phenomenon: across a large fraction of attention heads, Q and K vectors are highly concentrated around a fixed non-zero center, as shown in Figure~\ref{fig:intro-combined}(A)---a property we term \textit{Q/K concentration}. This concentration remains stable across positions and contexts, {\em i.e.}, Figure~\ref{fig:intro-combined}(C). Because pre-RoPE vectors---before positional encoding is applied---are unaffected by positional rotation, this stability is intrinsic rather than coincidental. Moreover, pre-RoPE vectors are directly linked to attention through the RoPE formula, making these centers meaningful for assessing KV importance.

%

To leverage these properties, we first demonstrate these centers relate to attention behavior via a trigonometric series. When Q/K are highly concentrated, they can be approximated by their centers; substituting these centers into the RoPE formula, the attention logit reduces to a function that depends only on Q-K distance---a trigonometric series---forming an attention-vs-distance curve. This curve usually exhibits peaks at specific Q-K distances, and verify that keys at those distances indeed receive higher attention in practice, as demonstrated in Figure~\ref{fig:intro-combined}(D). This demonstrates that Q/K concentration causes attention to favor keys at specific distances, with the centers determining which distances are preferred. Furthermore, this preference can be predicted using trigonometric series computed from centers.

We apply this understanding to design TriAttention, a KV cache compression method. TriAttention scores keys and retains only the top-scoring ones to address the memory bottleneck. The key idea of our scoring function is to use the Q center together with the trigonometric series to evaluate importance differences among keys arising from distance preferences. For the minority of heads where Q is less concentrated, we incorporate Q/K norms as complementary signals~\cite{guo2024attention,kobayashi2020attention,li2024survey}. We automatically balance these two components using a Q/K concentration metric.

We evaluate TriAttention on mathematical reasoning benchmarks. On AIME25, at equivalent accuracy to Full Attention, TriAttention achieves 2.5$\times$ higher throughput or 10.7$\times$ KV memory reduction (Figure~\ref{fig:teaser}), while R-KV achieves only about half the accuracy at the same efficiency. At fixed memory budget, the advantage is equally pronounced: TriAttention nearly doubles the accuracy of R-KV on AIME25 (32.9\% vs.\ 17.5\%) and AIME24 (42.1\% vs.\ 25.4\%). On MATH 500, with only 1,024 out of 32k tokens in the KV cache, TriAttention closely matches Full Attention (68.4\% vs.\ 69.6\%). These results demonstrate that leveraging Q/K centers via the trigonometric series provides a more reliable signal for KV importance than observation-based methods.

\begin{figure*}[t]
  \vskip 0.2in
  \begin{center}
    \includegraphics[width=0.95\textwidth]{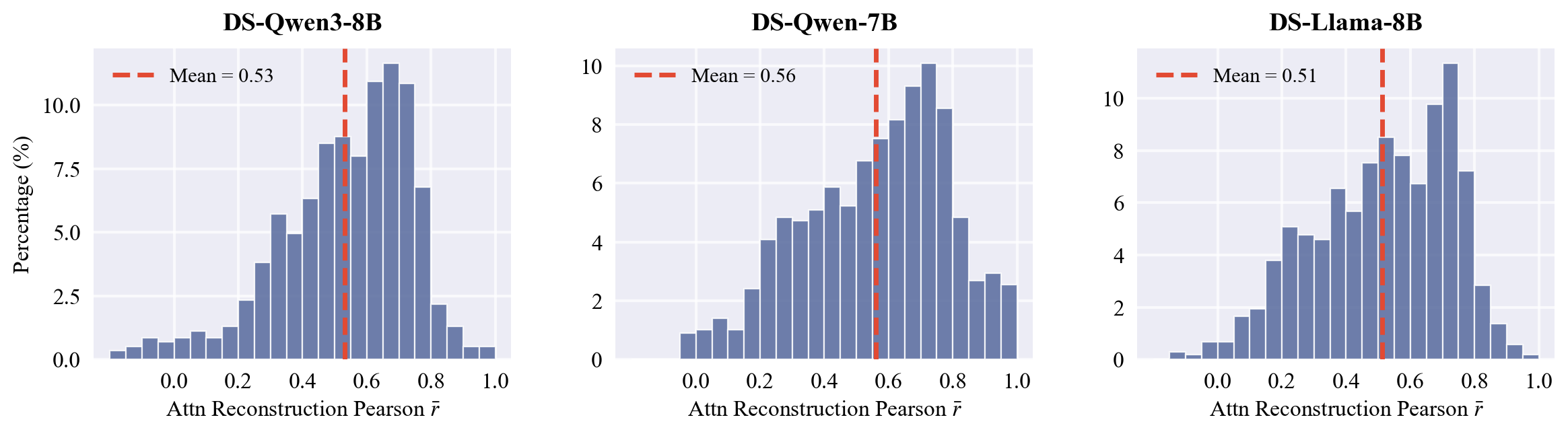}
    \caption{Attention reconstruction correlation across three DeepSeek-R1 distilled LLMs, including Qwen3~\cite{qwen2025qwen3}, Qwen2.5~\cite{qwen2024qwen25}, and Llama3~\cite{dubey2024llama3}. Distribution of per-head reconstruction Pearson correlation ($\bar{r}$) across all attention heads. The red dashed line indicates the mean. All models show right-skewed distributions with means above 0.5.}
    \label{fig:reconstruction-validation}
  \end{center}
  \vskip -0.2in
\end{figure*}
\section{Background and Related Work}



KV cache compression retains only a subset of KV pairs under a fixed memory budget. The core challenge is \textit{importance estimation}: determining which tokens will receive high attention from future queries. Existing methods approach this problem by analyzing post-RoPE representations, as we discuss in \S\ref{subsec:post-rope-methods}. To understand why post-RoPE analysis is limiting, we first review how RoPE works.

\subsection{Rotary Position Embedding}

Rotary Position Embedding (RoPE)~\cite{su2024roformer,hong2024token,barbero2025rope} encodes positional information as rotations in vector space and has become the dominant positional encoding in modern LLMs~\cite{dubey2024llama3,jiang2023mistral,qwen2025qwen3,peng2023yarn,team2025gemma}. For a $d$-dimensional vector, RoPE divides it into $d/2$ two-dimensional subspaces indexed by $f \in \{0, \ldots, d/2-1\}$. The $f$-th subspace rotates at frequency $\omega_f = \theta^{-2f/d}$ ($\theta = 10000$ typically); we refer to it as frequency band $f$. For frequency band $f$, RoPE applies a rotation by angle $\omega_f p$ at position $p$:
\begin{equation}
    \begin{pmatrix} x'_{2f} \\ x'_{2f+1} \end{pmatrix} =
    \begin{pmatrix} \cos(\omega_f p) & -\sin(\omega_f p) \\ \sin(\omega_f p) & \cos(\omega_f p) \end{pmatrix}
    \begin{pmatrix} x_{2f} \\ x_{2f+1} \end{pmatrix}
\end{equation}
We call the Q/K vectors before this rotation \textit{pre-RoPE}, and after rotation \textit{post-RoPE}. 

\subsection{Post-RoPE Compression Methods}
\label{subsec:post-rope-methods}

KV cache compression methods~\cite{cai2025rkv,ge2024model,feng2025adakv,behnam2025rocketkv} estimate token importance from post-RoPE representations. We categorize them into heuristic, attention-based, and norm-based types.

\textbf{Heuristic methods.}
Early compression methods use heuristic rule-based attention patterns.
StreamingLLM~\cite{xiao2024efficient,gu2025attentionsink} enables infinite-length streaming input
by retaining only a small fixed-size cache: a few initial ``sink'' tokens plus
a sliding window of recent tokens. The sink tokens exploit the observation that
initial positions receive disproportionate attention regardless of their content---discarding
them causes attention scores to lose a stable ``sink'' and degrades performance.
While simple, such fixed rules cannot adapt to content-dependent
importance, motivating more sophisticated approaches.

\textbf{Attention-based methods.}
These methods use attention scores---computed on post-RoPE Q/K---to identify important tokens.
H2O~\cite{zhang2023h2o} accumulates attention scores across decoding steps
to identify ``heavy-hitter'' tokens that consistently receive high attention.
SnapKV~\cite{li2024snapkv} computes attention within a local window
and aggregates scores to predict which tokens will matter for generation.
Scissorhands~\cite{liu2023scissorhands} exploits the ``persistence of importance''
hypothesis, using historical attention to guide eviction.
R-KV~\cite{cai2025rkv} scores tokens by attention from the most recent queries,
combined with redundancy detection for reasoning models.
LazyEviction~\cite{zhang2025lazyevictionlaggedkveviction} tracks token importance recurrence
within an observation window to delay eviction decisions.

\textbf{Norm-based methods.}
VATP~\cite{guo2024attention} observes that attention scores alone are insufficient:
attention sinks receive high attention but have near-zero value norms,
contributing little to the output. By incorporating value vector norms,
VATP provides a more nuanced importance measure.

\subsection{Limitations of Post-RoPE Methods}

Both attention-based and norm-based methods can be viewed as operating in the post-RoPE space, where positional rotations have already been applied. This shared foundation leads to distinct but related limitations.

For attention-based methods, the key constraint is that queries rotate with position, limiting useful observations to a tiny window and causing important keys undetected.

For norm-based methods, the limitation is different: they leverage only vector magnitudes while ignoring directional information. Ideally, incorporating Q/K directions would improve importance estimation---attention depends on both norms and the angle between Q and K. However, in post-RoPE space, directions are entangled with positional rotations: a vector's direction changes continuously with position, making directional information difficult to exploit.

\section{Pre-RoPE and Trigonometric Series}
\label{sec:observations}

As outlined in the Introduction, we observe that Q and K vectors in pre-RoPE space are highly concentrated around non-zero centers, and this concentration remains stable across positions and contexts. This is not a property of specific head types, but a prevalent phenomenon across models, as shown in Figure~\ref{fig:intro-combined}(A, C). In this section, we first characterize this concentration (\S\ref{subsec:concentration}), then show how it enables attention patterns to be described by a trigonometric series (\S\ref{subsec:mechanism}), and finally validate this through experiments (\S\ref{subsec:validation}).

\begin{figure*}[t]
  \vskip 0.2in
  \begin{center}
    \includegraphics[width=1.0\textwidth]{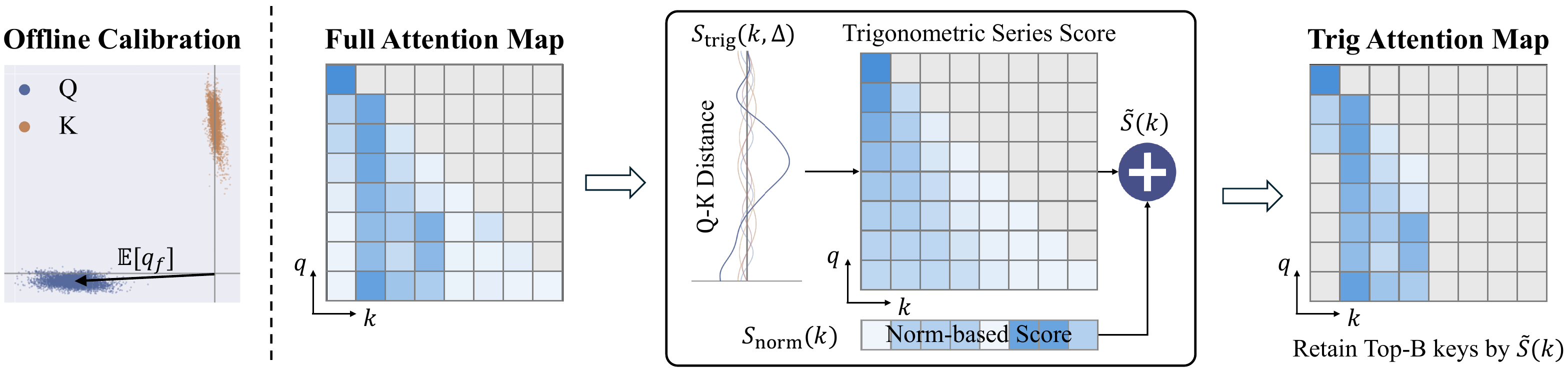}
    \caption{Method overview. From left to right: offline calibration computes Q distribution centers; then during inference, original attention is scored by combining $S_{\text{trig}}$ and norm-based components; the rightmost panel shows the attention map after pruning. We observe that some heads exhibit distance preference---distant keys tend to receive higher attention. However, we also find that certain keys, despite being far from the query, receive little attention due to their low norms. This motivates our two scoring components: $S_{\text{trig}}$ captures distance preference, while the norm-based score identifies low-norm keys. In this example, $S_{\text{trig}}$ correctly assigns low scores to nearby keys, while the norm-based score identifies the earliest token (leftmost) as unimportant due to its low norm, despite its maximal distance. Together, they accurately identify tokens that will not be attended to and prune them. See Appendix~\ref{app:method-visualization} for visualizations with real attention maps.}
    \label{fig:method-overview}
  \end{center}
  \vskip -0.2in
\end{figure*}
\subsection{The Pre-RoPE Concentration Phenomenon}
\label{subsec:concentration}

We examine Q/K distributions in the pre-RoPE space. For each head, we identify its \textit{dominant frequency bands}---the frequencies that contribute most to the attention logit~\cite{zhou2025elitekv} (see Appendix~\ref{app:dominant-bands})---and visualize Q/K vectors in the corresponding 2D planes.

\begin{observation}[Prevalent Q/K Concentration]
\label{obs:qk-concentration}
Q and K vectors in pre-RoPE space are highly concentrated around non-zero centers across most attention heads. This concentration is stable across different token positions and input contexts, as illustrated in Figure~\ref{fig:intro-combined}(A).
\end{observation}

To quantify this concentration, we use the \textbf{Mean Resultant Length} $R = \|\mathbb{E}[q]\| / \mathbb{E}[\|q\|]$, which measures how tightly vectors concentrate around their mean direction ($R \to 1$ indicates perfect concentration; $R \to 0$ indicates uniform dispersion). Figure~\ref{fig:intro-combined}(C) shows that across all heads in Qwen3-8B, the vast majority exhibit $R$ values approaching 1.0, confirming that Q/K concentration is prevalent.

This widespread concentration has important implications: when Q/K vectors are approximately constant, the attention computation simplifies dramatically, as we show next.

\subsection{Predictable Distance Preferences}
\label{subsec:mechanism}

When Q/K vectors are highly concentrated, we can approximate them by their centers. This approximation transforms the attention computation into a trigonometric series that depends only on Q-K distance, making attention patterns predictable from the centers alone.

Consider the RoPE attention formula. For a query $q$ at position $p_q$ and Key $k$ at position $p_k$, RoPE rotates frequency band $f$ at rate $\omega_f$. The pre-softmax logit is (see Appendix~\ref{app:rope-fourier}):
\begin{equation}
    \text{logit}(q, k) = \sum_{f} \|q_f\| \|k_f\| \cos(\omega_f \Delta + \phi_f)
    \label{eq:rope-logit}
\end{equation}
where $\Delta = p_q - p_k$ is the Q-K distance, $q_f, k_f \in \mathbb{C}$ are the pre-RoPE components in frequency band $f$, and $\phi_f = \arg(q_f) - \arg(k_f)$ is their phase difference.

When Q/K are concentrated, we approximate $q_f \approx \bar{q}_f$ and $k_f \approx \bar{k}_f$ (the centers). Since $\bar{q}_f$ and $\bar{k}_f$ are constants, the logit becomes a function of distance alone:
\begin{align}
    \text{logit}(\Delta) &\approx \sum_{f} \underbrace{\|\bar{q}_f\| \|\bar{k}_f\|}_{\text{amplitude}} \cos(\omega_f \Delta + \underbrace{\bar{\phi}_f}_{\text{phase}}) \nonumber \\
    &= \sum_{f} \left[ a_f \cos(\omega_f \Delta) + b_f \sin(\omega_f \Delta) \right]
    \label{eq:trig-series}
\end{align}
where coefficients $a_f, b_f$ are determined by the Q/K centers. This is a \textbf{trigonometric series} in Q-K distance $\Delta$.

Though RoPE frequencies follow a geometric rather than harmonic progression, the principle is analogous to Fourier synthesis: the learned Q/K centers determine the coefficients, which in turn shape the attention-vs-distance curve. Different centers produce different curves---some peak at small distances (local attention), others at large distances (attention sinks). In all cases, the distance preference is encoded in the Q/K centers and can be predicted via the trigonometric series.

\subsection{Experimental Validation}
\label{subsec:validation}

We experimentally test whether Q/K concentration causes attention to follow the distance preferences described by the trigonometric series. We compute the series from Q/K centers and check if it reconstructs actual attention. Successful reconstruction confirms this causal link and shows these preferences are predictable from the centers.

We test this on Qwen3-8B across all 1152 attention heads (36 layers $\times$ 32 heads) using a $\sim$10K token sequence. For reconstruction, we compute the mean Q and K vectors from a calibration dataset in the pre-RoPE space---denoted $\mathbb{E}[q_f]$ and $\mathbb{E}[k_f]$ for frequency band $f$---and substitute them into the trigonometric series:
\begin{equation}
    \hat{s}(\Delta) = \sum_{f} \|\mathbb{E}[q_f]\| \|\mathbb{E}[k_f]\| \cos(\omega_f \Delta + \phi_f)
    \label{eq:reconstruction}
\end{equation}
where $\phi_f = \arg(\mathbb{E}[q_f]) - \arg(\mathbb{E}[k_f])$ is the phase difference between the mean vectors. This yields a predicted attention curve over Q-K distance $\Delta$ (Figure~\ref{fig:intro-combined}(D)).

To quantify prediction quality, we define the \textbf{Reconstruction Correlation $\bar{r}$}: the mean Pearson correlation between predicted and actual attention logits. For each query, we compute the correlation between its actual logits and the predicted curve, then average across all queries:
\begin{equation}
    \bar{r} = \frac{1}{N} \sum_{i=1}^{N} \rho(\mathbf{a}_i, \hat{\mathbf{s}})
\end{equation}
Here $\mathbf{a}_i$ is the vector of actual attention logits for query $i$, $\hat{\mathbf{s}}$ is the predicted logits from Equation~\ref{eq:reconstruction}, and $\rho$ is Pearson correlation. Both $\mathbf{a}_i$ and $\hat{\mathbf{s}}$ are evaluated at the same logarithmically-spaced distances $\Delta = 1, 2, 4, 8, \ldots$, ensuring balanced coverage across distance scales.

Figure~\ref{fig:intro-combined}(D) shows an example. For the first head of the first layer---chosen to avoid cherry-picking---the prediction closely tracks actual attention, achieving $\bar{r} = 0.72$. Across all heads in three different architectures (Qwen3, Qwen2.5, Llama3), $\bar{r}$ peaks around 0.6--0.9 in the distribution, with mean values above 0.5; see Figure~\ref{fig:reconstruction-validation} for the full distributions. The high correlation across many heads and architectures confirms that the trigonometric series computed from Q/K centers accurately predicts attention patterns. We further find that Q/K concentration is a model-intrinsic property: on Qwen3-8B, measuring MRL across Math, Coding, and Chat domains yields nearly identical values (0.977--0.980), with ${\sim}90\%$ of heads exhibiting $R > 0.95$ regardless of domain. The same holds across architectures, including MLA (Appendix~\ref{app:mla}).

\section{TriAttention}
\label{sec:method}

Based on the analysis in \S\ref{sec:observations}, we propose a KV cache compression method that scores key importance and retains only the top-scoring keys (Figure~\ref{fig:method-overview}). The scoring function combines two signals (\S\ref{subsec:scoring-functions}): the trigonometric series, which captures distance preferences, and norm information as a complement; the weight between them is adjusted based on Q/K concentration (\S\ref{subsec:balancing}). Finally, we describe how to apply this scoring for KV cache pruning (\S\ref{subsec:pruning}).

\begin{table*}[t]
    \caption{Reasoning performance on AIME24 and AIME25. Best results are in \textbf{bold}, second best are \underline{underlined}. All methods are compared under the same KV cache budget.}
    \centering
    \resizebox{\textwidth}{!}{
    \begin{tabular}{l|cccc|cccc}
        \toprule
        \multirow{2}{*}{\textbf{Method}} & \multicolumn{4}{c|}{\textbf{AIME24}} & \multicolumn{4}{c}{\textbf{AIME25}} \\
        \cmidrule(lr){2-5} \cmidrule(lr){6-9}
        & Qwen3-8B & DS-Llama & DS-Qwen & GPT-OSS & Qwen3-8B & DS-Llama & DS-Qwen & GPT-OSS \\
        \midrule
        Full Attention & 57.1 & 50.4 & 43.8 & 69.2 & 40.8 & 31.4 & 34.2 & 60.0 \\
        \midrule
        SnapKV & \underline{34.6} & 5.0 & \underline{34.6} & 48.3 & \underline{20.0} & 6.7 & \underline{25.0} & 36.7 \\
        R-KV & 25.4 & \underline{25.8} & \underline{34.6} & \underline{49.6} & 17.5 & \underline{11.2} & 23.3 & \underline{39.2} \\
        \rowcolor{lm_purple} TriAttention & \textbf{42.1} & \textbf{33.8} & \textbf{42.5} & \textbf{59.2} & \textbf{32.9} & \textbf{19.6} & \textbf{30.0} & \textbf{49.2} \\
        \bottomrule
    \end{tabular}
    }
    \label{tab:reasoning_aime}
\end{table*}

\subsection{Key Importance Scoring}
\label{subsec:scoring-functions}

In KV cache, keys are already cached. Let $k_f \in \mathbb{C}$ denote the pre-RoPE of a key $k$ in frequency band $f$, and $\Delta = p_q - p_k$ the Q-K distance.

\textbf{Trigonometric Series Score.} Keys at different positions have different distances to future queries. To estimate how much attention each key will receive, we use the Q center as a proxy for future queries---justified by the Q/K concentration observed in \S\ref{sec:observations}. The trigonometric series then gives the expected attention at distance $\Delta$:
\begin{equation}
    S_{\text{trig}}(k, \Delta) = \sum_{f} \|\mathbb{E}[q_f]\| \cdot \|k_f\| \cdot \cos(\omega_f \Delta + \phi_f)
\end{equation}
Here $\mathbb{E}[q_f] \in \mathbb{C}$ is the Q center in frequency band $f$, computed from calibration data. Since keys in the cache are known, we directly use their representations $k_f$. The phase $\phi_f = \arg(\mathbb{E}[q_f]) - \arg(k_f)$ is the angular difference between the Q center and the current key.

\textbf{Norm-Based Score.} The trigonometric series assumes Q/K are exactly at their centers. In practice, there is variation around the centers, and this variation affects attention. We account for this with a norm-based term:
\begin{equation}
    S_{\text{norm}}^{(0)}(k) = \sum_{f} \mathbb{E}[\|q_f\|] \cdot \|k_f\|
\end{equation}
Here $\mathbb{E}[\|q_f\|]$ is the expected query norm in band $f$, computed from calibration data. Unlike methods that only consider $\|k\|$, we weight each frequency band by its expected query contribution.

\subsection{Adaptive Weighting via Concentration}
\label{subsec:balancing}

The trigonometric series score $S_{\text{trig}}$ is most accurate when Q/K are highly concentrated; the norm-based score $S_{\text{norm}}$ becomes more important for the few heads where concentration is lower. We automatically balance these components using Q/K concentration as a weighting factor.

Recall that concentration is quantified by the \textbf{Mean Resultant Length} $R_f = \|\mathbb{E}[q_f]\| / \mathbb{E}[\|q_f\|]$ for each frequency band $f$. When $R_f$ is high, the trigonometric series is accurate and $S_{\text{trig}}$ is reliable; when $R_f$ is low, there is more variation around the center and $S_{\text{norm}}$ provides useful complementary information.

We refine the norm-based score by scaling each frequency band by $(1 - R_f)$:
\begin{equation}
    S_{\text{norm}}(k) = \sum_{f} (1 - R_f) \cdot \mathbb{E}[\|q_f\|] \cdot \|k_f\|
\end{equation}
This can be rewritten as:
\begin{equation}
    S_{\text{norm}}(k) = \sum_{f} \left(\mathbb{E}[\|q_f\|] - \|\mathbb{E}[q_f]\|\right) \cdot \|k_f\|
\end{equation}
Intuitively, when $R_f$ is high (concentration is strong), $(1-R_f)$ is small, so $S_{\text{norm}}$ contributes little and $S_{\text{trig}}$ dominates. When $R_f$ is low (concentration is weak), the full norm contribution is preserved.

The final combined score is:
\begin{equation}
    S(k, \Delta) = S_{\text{trig}}(k, \Delta) + S_{\text{norm}}(k)
\end{equation}
A key may be queried from any future position, so its importance depends on all future query positions. We compute $S(k, \Delta + \delta)$ at multiple offsets and define the averaged importance score as:
\begin{equation}
    \tilde{S}(k) = \frac{1}{|\mathcal{D}|} \sum_{\delta \in \mathcal{D}} S(k, \Delta + \delta)
\end{equation}
where $\mathcal{D} = \{1, 2, 4, \ldots, 2^{16}\}$ is the set of future offsets.

\subsection{KV Cache Pruning}
\label{subsec:pruning}

With the scoring function $\tilde{S}(k)$ defined, we score each key in the cache independently and retain only the top-$B$. Before selection, we address two practical considerations: reducing the computational overhead of frequent scoring, and handling Grouped-Query Attention (GQA)~\cite{ainslie2023gqa} where multiple query heads share each KV head.

\textbf{Window-based Pruning.}
Scoring all keys at every decoding step is computationally expensive. Instead, we trigger pruning once every $\beta = 128$ generated tokens: when the 128-th token of each interval is generated, if the cache exceeds budget $B$, we score all keys, retain the top-$B$, and evict the rest. We call each 128-token interval a \textit{window}, following R-KV~\cite{cai2025rkv}. This batched approach significantly reduces overhead.

\textbf{Grouped-Query Attention.}
In GQA, each KV head is shared by $G$ query heads. Since $\tilde{S}$ depends on query statistics ($\mathbb{E}[q_f]$ and $\mathbb{E}[\|q_f\|]$), each key receives $G$ different scores. These scores operate at different scales across heads, making direct comparison unreliable.

We apply normalize-then-aggregate. Let $\tilde{S}^{(g)}(k)$ denote the score computed using statistics from query head $g$. We first z-score normalize within each head:
\begin{equation}
    \hat{S}^{(g)}(k) = \frac{\tilde{S}^{(g)}(k) - \mu_g}{\sigma_g}
\end{equation}
where $\mu_g, \sigma_g$ are the mean and standard deviation of $\{\tilde{S}^{(g)}(k)\}_k$ over all keys. We then aggregate via maximum:
\begin{equation}
    S_{\text{final}}(k) = \max_{g \in \{0, \ldots, G-1\}} \left( \hat{S}^{(g)}(k) \right)
\end{equation}
A key is retained if \textit{any} query head deems it important. We retain the top-$B$ keys by $S_{\text{final}}(k)$ and evict the rest.

\begin{figure*}[t]
  \vskip 0.2in
  \begin{center}
    \includegraphics[width=0.95\textwidth]{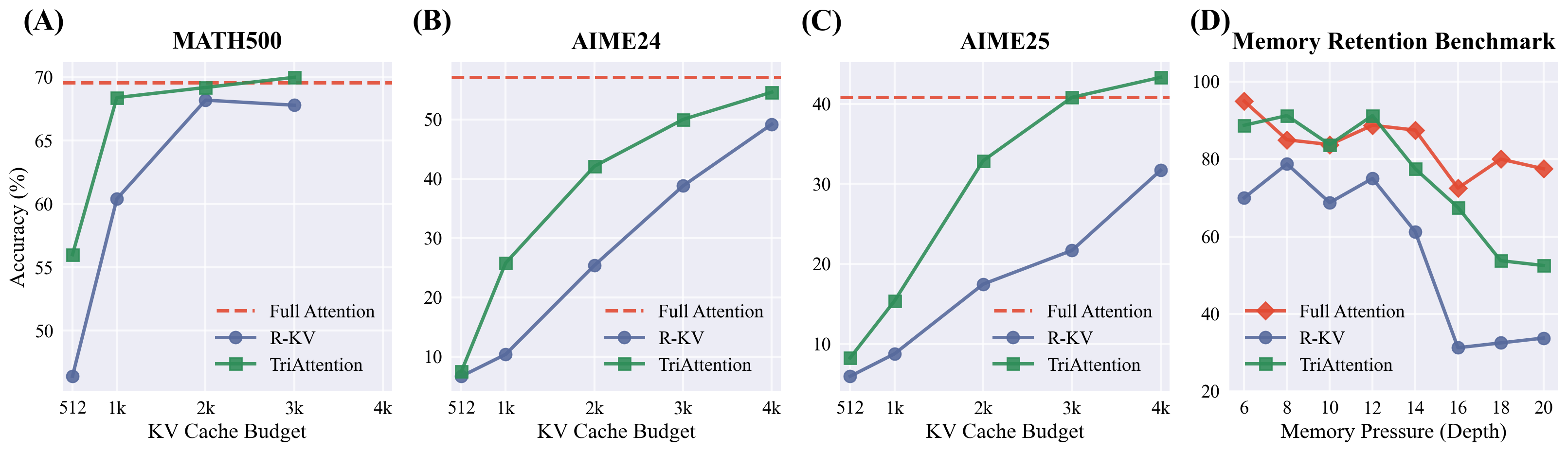}
    \caption{Performance comparison on Qwen3-8B. \textbf{(A--C)} Accuracy vs.\ KV cache budget on three mathematical reasoning benchmarks. TriAttention consistently outperforms R-KV across all budget levels. \textbf{(D)} Memory retention on Recursive State Query benchmark. Depth refers to DFS recursion depth; deeper recursion requires retaining more intermediate states, increasing memory pressure.}
    \label{fig:budget_curves}
  \end{center}
  \vskip -0.2in
\end{figure*}
\section{Experiments}
\label{sec:experiments}

\subsection{Experimental Setup}
\label{subsec:setup}

\textbf{Models.}
We evaluate on four reasoning-capable LLMs spanning different architectures and scales: Qwen3-8B~\cite{qwen2025qwen3}, DeepSeek-R1-Distill-Llama-8B~\cite{deepseek2025r1}, DeepSeek-R1-Distill-Qwen-7B~\cite{deepseek2025r1}, and GPT-OSS-20B~\cite{openai2025gptoss}. These models are selected for their strong chain-of-thought reasoning capabilities and diverse architectural foundations.

\textbf{Baselines.}
We compare against three methods: (1) \textit{Full Attention}---no pruning, serving as the performance upper bound; (2) \textit{SnapKV}~\cite{li2024snapkv}---selects important tokens based on historical attention scores from a local observation window; (3) \textit{R-KV}~\cite{cai2025rkv}---a recent method combining attention-based importance scoring with redundancy detection for reasoning models. Additional comparisons with other methods are in Appendix~\ref{app:general-tasks}.

\textbf{Datasets.}
We use three mathematical reasoning benchmarks: AIME 2024~\cite{aime2024} (30 problems), AIME 2025~\cite{aime2025} (30 problems), and MATH 500~\cite{hendrycksmath2021} (500 problems). AIME problems are challenging competition-level mathematics requiring multi-step reasoning; MATH 500 covers diverse mathematical reasoning tasks. We further evaluate on general tasks (LongBench, RULER) in Appendix~\ref{app:general-tasks}.

\begin{table}[t]
    \caption{Reasoning performance on MATH 500 with KV budget of 512. Best results are in \textbf{bold}, second best are \underline{underlined}.}
    \centering
    \resizebox{\columnwidth}{!}{
    \begin{tabular}{l|cccc}
        \toprule
        \textbf{Method} & Qwen3-8B & DS-Llama & DS-Qwen & GPT-OSS \\
        \midrule
        Full Attention & 69.6 & 82.4 & 87.0 & 91.4 \\
        \midrule
        SnapKV & \underline{49.2} & 65.5 & 66.4 & 68.2 \\
        R-KV & 46.4 & \underline{76.9} & \underline{71.6} & \underline{77.4} \\
        \rowcolor{lm_purple} TriAttention & \textbf{56.0} & \textbf{80.6} & \textbf{79.6} & \textbf{81.2} \\
        \bottomrule
    \end{tabular}
    }
    \label{tab:reasoning_math500}
\end{table}
\textbf{Implementation Details.}
Unless otherwise specified, experiments are conducted on NVIDIA A100 80GB GPUs using bfloat16 precision with FlashAttention-2~\cite{dao2024flashattention}. GPT-OSS uses H100 GPUs with FlashAttention-3~\cite{shah2024flashattention}, as FlashAttention-3 requires Hopper architecture. For generation, we set maximum length to 32,768 (32k) tokens, temperature to 0.6, and top-p to 0.95. For AIME benchmarks, each problem is sampled 8 times and we report the average pass rate; for MATH 500, we sample once per problem. For KV cache compression, we follow the same settings as R-KV: every 128 tokens decoded, a compression round is triggered to prune the KV cache back to the specified budget. The default KV budget is 2048 tokens. For DS-Llama and MATH 500, we use a budget of 512 tokens due to shorter response lengths, ensuring that KV cache compression is actually exercised during generation.

\subsection{Results}
\label{subsec:results}

\subsubsection{Reasoning Task Performance}
\label{subsubsec:reasoning}

We evaluate TriAttention on standard mathematical reasoning benchmarks. Tables~\ref{tab:reasoning_aime} and~\ref{tab:reasoning_math500} present results on AIME and MATH 500 respectively.
Across all models and datasets, TriAttention consistently achieves the best performance among KV cache compression methods, closely approaching Full Attention while outperforming all baselines.

To understand how performance varies with compression aggressiveness, we evaluate TriAttention and R-KV on Qwen3-8B across a range of KV budgets from 512 to 4096 tokens on three benchmarks, as shown in Figure~\ref{fig:budget_curves}. TriAttention consistently outperforms R-KV across all budget levels, with the advantage being most pronounced at low-to-mid budgets. On MATH 500, TriAttention matches Full Attention at a budget of 1024 and slightly exceeds it at higher budgets. On the more challenging AIME benchmarks, TriAttention maintains a substantial lead: with a budget of 2048 on AIME25, TriAttention achieves 32.9\% compared to R-KV's 17.5\%---a 15.4 percentage point gap. Notably, TriAttention matches or exceeds Full Attention with significantly compressed KV cache: on AIME25, it achieves 43.3\% with a budget of 4096, surpassing Full Attention.

\begin{table*}[t]
    \caption{Ablation studies on Qwen3-8B with KV budget of 2048. (A) Effect of removing trigonometric series score $S_{\text{trig}}$. (B) Effect of removing concentration-based weighting. (C) Cross-domain generalization: comparing calibration on coding vs.\ reasoning data, both tested on reasoning.}
    \centering
    \begin{subtable}[t]{0.30\textwidth}
        \centering
        \caption{Trigonometric series score}
        \begin{tabular}{l|cc}
            \toprule
            \textbf{Method} & \textbf{AIME24} & \textbf{AIME25} \\
            \midrule
            w/o $S_{\text{trig}}$ & 18.8 & 21.2 \\
            \rowcolor{lm_purple} TriAttn. & 42.1 & 32.9 \\
            \bottomrule
        \end{tabular}
        \label{tab:ablation_trig}
    \end{subtable}
    \hfill
    \begin{subtable}[t]{0.30\textwidth}
        \centering
        \caption{Concentration-based weighting}
        \begin{tabular}{l|cc}
            \toprule
            \textbf{Method} & \textbf{AIME24} & \textbf{AIME25} \\
            \midrule
            w/o $R$ & 41.3 & 28.7 \\
            \rowcolor{lm_purple} TriAttn. & 42.1 & 32.9 \\
            \bottomrule
        \end{tabular}
        \label{tab:ablation_r}
    \end{subtable}
    \hfill
    \begin{subtable}[t]{0.30\textwidth}
        \centering
        \caption{Cross-domain calibration}
        \begin{tabular}{l|cc}
            \toprule
            \textbf{Method} & \textbf{AIME24} & \textbf{AIME25} \\
            \midrule
            Coding & \textbf{44.2} & 29.2 \\
            Reasoning & 42.1 & \textbf{32.9} \\
            \bottomrule
        \end{tabular}
        \label{tab:ablation_calib}
    \end{subtable}
    \label{tab:ablation}
\end{table*}

\begin{table*}[t]
    \caption{Throughput comparison on Qwen3-8B. At KV budgets achieving comparable accuracy to Full Attention, TriAttention shows significant throughput gains.}
    \centering
    \begin{tabular}{l|cc|cc|cc}
        \toprule
        \multirow{2}{*}{\textbf{Metric}} & \multicolumn{2}{c|}{\textbf{MATH 500}} & \multicolumn{2}{c|}{\textbf{AIME24}} & \multicolumn{2}{c}{\textbf{AIME25}} \\
        \cmidrule(lr){2-3} \cmidrule(lr){4-5} \cmidrule(lr){6-7}
        & \textbf{Full Attn.} & \cellcolor{lm_purple}\textbf{TriAttn.} & \textbf{Full Attn.} & \cellcolor{lm_purple}\textbf{TriAttn.} & \textbf{Full Attn.} & \cellcolor{lm_purple}\textbf{TriAttn.} \\
        \midrule
        KV Budget & -- & \cellcolor{lm_purple}1024 & -- & \cellcolor{lm_purple}4096 & -- & \cellcolor{lm_purple}3072 \\
        Accuracy & 69.6 & \cellcolor{lm_purple}68.4 & 57.1 & \cellcolor{lm_purple}54.6 & 40.8 & \cellcolor{lm_purple}40.8 \\
        Throughput & 222.8 & \cellcolor{lm_purple}\textbf{1405.2} & 222.8 & \cellcolor{lm_purple}\textbf{413.9} & 222.8 & \cellcolor{lm_purple}\textbf{563.5} \\
        \midrule
        \cellcolor{lm_purple}Speedup & \multicolumn{2}{c|}{\cellcolor{lm_purple}\textbf{6.3$\times$}} & \multicolumn{2}{c|}{\cellcolor{lm_purple}\textbf{1.9$\times$}} & \multicolumn{2}{c}{\cellcolor{lm_purple}\textbf{2.5$\times$}} \\
        \bottomrule
    \end{tabular}
    \label{tab:fullkv_comparison}
\end{table*}

\begin{table}[t]
    \caption{Comparison between R-KV and TriAttention at comparable reasoning accuracy and memory.}
    \centering
    \resizebox{\columnwidth}{!}{
    \begin{tabular}{l|cc|cc}
        \toprule
        \multirow{2}{*}{\textbf{Metric}} & \multicolumn{2}{c|}{\textbf{Comparable Accuracy}} & \multicolumn{2}{c}{\textbf{Comparable Memory}} \\
        \cmidrule(lr){2-3} \cmidrule(lr){4-5}
        & \textbf{R-KV} & \cellcolor{lm_purple}\textbf{TriAttn.} & \textbf{R-KV} & \cellcolor{lm_purple}\textbf{TriAttn.} \\
        \midrule
        KV Budget & 2048 & \cellcolor{lm_purple}1024 & 1024 & \cellcolor{lm_purple}1024 \\
        MATH 500 & 68.2 & \cellcolor{lm_purple}\textbf{68.4} & 60.4 & \cellcolor{lm_purple}\textbf{68.4} $_{\textcolor{red}{\textcolor{red}{\uparrow 8.0}}}$ \\
        AIME24 & 25.4 & \cellcolor{lm_purple}\textbf{25.8} & 10.4 & \cellcolor{lm_purple}\textbf{25.8} $_{\textcolor{red}{\textcolor{red}{\uparrow 15.4}}}$ \\
        Throughput & 760.4 & \cellcolor{lm_purple}\textbf{1405.2} $_{\textcolor{red}{\textcolor{red}{\uparrow 644.8}8}}$ & 1345.5 & \cellcolor{lm_purple}\textbf{1405.2} \\
        \bottomrule
    \end{tabular}
    }
    \label{tab:rkv_comparison_overall}
\end{table}

\subsubsection{Memory Retention Benchmark}
\label{subsubsec:memory}

KV cache pruning may cause LLMs to forget intermediate states during long reasoning chains. To measure this effect, we design a benchmark based on \textit{recursive simulation}.

\textbf{Why Recursion Tests Memory.}
Recursive algorithms naturally stress memory retention: the model must descend through nested calls, retain intermediate states, then backtrack to produce results. Deeper recursion depth requires retaining more intermediate states, creating increasing memory pressure. If a KV cache pruning algorithm incorrectly discards these critical states, the model loses the information needed for backtracking, and errors cascade through all subsequent steps (see Appendix~\ref{app:dfs-benchmark} for an illustration). We use depth-first search (DFS)~\cite{reif1985depth} to instantiate this recursive simulation; see Appendix~\ref{app:dfs-benchmark} for task details.

We evaluate on Qwen3-8B with step counts from 6 to 20, comparing Full Attention, R-KV, and TriAttention on 80 samples per step count. Both compression methods use a KV budget of 2048, as shown in Figure~\ref{fig:budget_curves}(D).

Under low-to-moderate memory pressure up to depth 16, TriAttention performs comparably to Full Attention, even slightly outperforming it at depths 8 and 12. Only beyond depth 18 does TriAttention begin to lag behind. This suggests that Full Attention's KV cache contains redundancy under typical workloads, and TriAttention effectively preserves the essential information for memory retention.

In contrast, R-KV consistently underperforms both methods across all depths. Most notably, R-KV exhibits catastrophic accuracy degradation starting at depth 16, dropping from approximately 61\% at depth 14 to 31\% at depth 16. This sharp decline indicates that the pruning of R-KV significantly impairs the model of retaining intermediate states.

\subsubsection{Ablation Study}
\label{subsubsec:ablation}


\textbf{Effect of Scoring Components.} Our scoring function combines the trigonometric series score $S_{\text{trig}}$ with the norm-based score $S_{\text{norm}}$. We validate both components by removing each in turn. Removing $S_{\text{trig}}$ and relying solely on $S_{\text{norm}}$ substantially degrades performance (Table~\ref{tab:ablation_trig}), confirming that distance preferences from the trigonometric series are essential. Conversely, removing $S_{\text{norm}}$ and relying solely on $S_{\text{trig}}$ drops AIME24 accuracy from 45.8\% to 40.4\% ($-5.4\%$), confirming that the norm-based score provides complementary information about token salience.

\textbf{Effect of Adaptive Weighting.} The Mean Resultant Length $R$ balances the two scoring components based on Q/K concentration. We compare against a variant ``w/o $R$'' that replaces $S_{\text{norm}}$ with $S_{\text{norm}}^{(0)}$, removing the $(1-R_f)$ weighting. Table~\ref{tab:ablation_r} shows that removing $R$ consistently degrades performance, validating the effectiveness of using Q/K concentration to adaptively weight the scoring components.

\textbf{Cross-Domain Calibration.} TriAttention requires offline calibration to collect Q/K statistics. A natural concern is whether these statistics overfit to the calibration domain. To test generalization, we calibrate on coding data~\cite{jain2025livecodebench} instead of reasoning data and evaluate on reasoning benchmarks (``Coding'' in Table~\ref{tab:ablation_calib}). Table~\ref{tab:ablation_calib} shows that cross-domain calibration achieves comparable accuracies, demonstrating that the learned Q/K statistics generalize across domains rather than overfitting to specific tasks.

\textbf{Future Offset Design.} Our scoring function evaluates key importance at multiple future offsets $\mathcal{D}$. We ablate two aspects: the range/number of offsets, and the spacing strategy (Table~\ref{tab:ablation_offset}). Increasing the maximum distance from 128 to 4096 improves accuracy from 41.7\% to 48.8\% (+7.1\%), confirming that future offsets are beneficial. Geometric spacing ($\{1, 2, 4, \ldots\}$) dramatically outperforms linear spacing (45.8\% vs.\ 28.7\%, $-17.1\%$), as near-distance positions require denser sampling.

\textbf{Calibration Sensitivity.} We test robustness to calibration data quantity and quality (Table~\ref{tab:ablation_calib_sensitivity}). Performance is stable across calibration sizes from 50k to 960k tokens (45.4--45.8\%). Similarly, calibration data quality shows no clear correlation with accuracy: using Google homepage HTML (low quality) achieves 46.2\%, comparable to ShareGPT chat data (46.7\%). This confirms that the Q/K statistics captured during calibration are model-intrinsic properties, robust to the choice of calibration data.

\subsubsection{Throughput and Efficiency}
\label{subsubsec:throughput}

We evaluate efficiency from two perspectives: comparison with Full Attention and with prior compression methods. Following the evaluation protocol of R-KV~\cite{cai2025rkv}, throughput is measured as the average tokens generated per second over 16K decoding length on a single A100 80GB GPU at maximum batch size.

\textbf{Comparison with Full Attention.} At comparable accuracy, TriAttention achieves substantial efficiency gains (Figure~\ref{fig:teaser}, Table~\ref{tab:fullkv_comparison}). On AIME25, at identical accuracy to Full Attention (40.8\%), TriAttention achieves 2.5$\times$ higher throughput or 10.7$\times$ KV memory reduction. On MATH 500, the throughput gain reaches 6.3$\times$ (1,405 vs.\ 223 tokens/s) while maintaining comparable accuracy (68.4\% vs.\ 69.6\%).

\textbf{Comparison with R-KV.} Table~\ref{tab:rkv_comparison_overall} compares TriAttention with R-KV under two settings. At comparable accuracy, TriAttention requires only half the KV budget (1,024 vs.\ 2,048 tokens) while achieving 85\% higher throughput (1,405 vs.\ 760 tokens/s). At the same memory budget, TriAttention maintains similar throughput but achieves higher accuracy---8\% on MATH 500 and 15\% on AIME24. TriAttention provides a strictly better accuracy-efficiency trade-off.

\section{Conclusion}

We discover \textit{Q/K concentration} in the pre-RoPE space: query and key vectors cluster around fixed centers, causing attention to follow predictable distance preferences via a trigonometric series. This concentration is a model-intrinsic property, consistent across domains and architectures. TriAttention leverages this finding to estimate key importance from stable Q/K centers, avoiding the instability of post-RoPE methods. Experiments on both mathematical reasoning (AIME, MATH 500) and general tasks (LongBench, RULER) show TriAttention consistently outperforms all baselines, matching Full Attention while achieving 2.5$\times$ throughput and 10.7$\times$ KV memory reduction.

\section*{Impact Statement}
This paper advances inference-time reasoning efficiency in LLMs. The proposed TriAttention cuts computational cost and latency without compromising accuracy, enabling more efficient, accessible deployment of reasoning-capable LLMs. Its societal and ethical implications align with standard ML efficiency improvements—including sustainability and usability benefits—and it introduces no new model capabilities, data sources, or application domains, thus raising no novel ethical concerns beyond those already established for LLMs, with no further broader impacts.

\section*{Acknowledgements}
We thank MIT-IBM Watson AI Lab, Amazon, Dell, Hyundai Motor Company, MIT AI Hardware Program for supporting this research.

\bibliography{main}

\bibliographystyle{icml2026}

\newpage
\appendix
\onecolumn

\section*{Appendix}
\setcounter{figure}{0}
\setcounter{table}{0}
\renewcommand{\thefigure}{\Alph{figure}} 
\renewcommand{\thetable}{\Alph{table}} 

\section{Limitations and Future Work}

In this work, we presented TriAttention, a novel KV cache compression mechanism that exploits the intrinsic stability of Q/K concentration in the pre-RoPE space to estimate key importance through trigonometric series. Our extensive evaluation on complex reasoning benchmarks confirms that TriAttention effectively matches the accuracy of Full Attention, while simultaneously achieving a $2.5\times$ increase in throughput and a $10.7\times$ reduction in KV memory footprint. While the current implementation demonstrates substantial efficiency gains, further latency reductions can be realized through the design of a dedicated, hardware-aware inference kernel optimized for our unique compression operations. A primary avenue for future improvement involves the development of a dedicated, high-performance inference kernel designed to further accelerate the computation of trigonometric series and the subsequent cache pruning process. Beyond these engineering optimizations, we plan to extend our evaluation to broader domains such as coding and agentic tasks, and to investigate refined compression strategies, including head-specific budgets, to push the boundaries of efficiency even further.



\section{RoPE and Trigonometric Series Connection}
\label{app:rope-fourier}

In this appendix, we establish the mathematical connection between RoPE (Rotary Position Embedding) and trigonometric series, explaining how attention heads can leverage RoPE to achieve distance-dependent attention patterns.

\subsection{RoPE as Complex Rotation}

RoPE applies position-dependent rotations to Query and Key vectors. For a 2D subspace corresponding to frequency $\omega_f$, RoPE can be written in complex form:
\begin{equation}
    \tilde{q}_f(p) = q_f \cdot e^{i \omega_f p}, \quad \tilde{k}_f(p) = k_f \cdot e^{i \omega_f p}
\end{equation}
where $q_f, k_f \in \mathbb{C}$ are the complex representations of Query and Key in frequency band $f$, and $p$ is the position.

\subsection{General Form of RoPE Attention}

The dot product between Query at position $p_q$ and Key at position $p_k$ in frequency band $f$ is:
\begin{equation}
    \text{Re}(\tilde{q}_f(p_q) \cdot \overline{\tilde{k}_f(p_k)}) = \text{Re}(q_f \overline{k_f} \cdot e^{i \omega_f (p_q - p_k)})
\end{equation}

Summing over all frequency bands, the attention logit is:
\begin{equation}
    \langle q, k \rangle_\Delta = \sum_{f} \|q_f\| \|k_f\| \cos(\omega_f \Delta + \phi_f)
\end{equation}
where $\Delta = p_q - p_k$ is the relative position, and $\phi_f = \arg(q_f) - \arg(k_f)$ is the phase difference. \textbf{Crucially}, the coefficients $\|q_f\|\|k_f\|$ and phases $\phi_f$ depend on the specific Q/K vectors, which vary across different tokens.

\subsection{Special Case: Constant Q/K Yields Trigonometric Series}

When pre-RoPE Q and K vectors are \textbf{constant} across tokens, the coefficients become fixed. Using the angle addition formula, the logit reduces to:
\begin{equation}
    \langle q, k \rangle_\Delta = \sum_{f} \left[ a_f \cos(\omega_f \Delta) + b_f \sin(\omega_f \Delta) \right]
    \label{eq:fourier-attn}
\end{equation}
where the coefficients are now constants:
\begin{align}
    a_f &= \|q_f\| \|k_f\| \cos(\phi_f) \\
    b_f &= -\|q_f\| \|k_f\| \sin(\phi_f)
\end{align}

This is a \textbf{trigonometric series} in $\Delta$. We note that RoPE frequencies follow a \textbf{geometric progression} $\omega_f = \theta^{-2f/d}$ (where $\theta = 10000$), not the harmonic progression $\omega_n = n\omega_0$ of classical Fourier series.

The key insight is that:
\begin{itemize}
    \item The \textbf{frequencies} $\omega_f$ are predetermined by RoPE (geometric progression)
    \item When Q/K are constant, the \textbf{coefficients} $(a_f, b_f)$ become fixed constants
    \item The model can ``synthesize'' arbitrary distance-dependent attention patterns by learning appropriate Q/K values
\end{itemize}

\subsection{From Q/K Concentration to Predictable Distance Preferences}

When Q/K vectors are highly concentrated around their centers---as quantified by high Mean Resultant Length $R$---the expected attention logit can be approximated using these centers:
\begin{equation}
    \mathbb{E}[\langle q, k \rangle_\Delta] \approx \sum_{f} \|\mathbb{E}[q_f]\| \|\mathbb{E}[k_f]\| \cos(\omega_f \Delta + \phi_f)
\end{equation}
where $\phi_f = \arg(\mathbb{E}[q_f]) - \arg(\mathbb{E}[k_f])$ is the phase difference between the mean vectors.

This approximation becomes accurate when concentration is high (i.e., $R \to 1$). In this regime, the Q/K centers fully determine the distance preference curve: different centers produce different attention-vs-distance curves, with peaks at specific Q-K distances. Crucially, these preferences are \textbf{predictable} from the centers alone, without observing actual attention scores---this is the key insight exploited by TriAttention.

\subsection{Mean Resultant Length}
\label{app:mean-resultant-length}

The Mean Resultant Length $R$ is a standard measure from directional statistics~\cite{mardia1999directional} that quantifies how tightly a distribution of vectors concentrates around its mean direction. For a set of unit vectors $\{u_1, \ldots, u_n\}$, the mean resultant length is defined as:
\begin{equation}
    R = \left\| \frac{1}{n} \sum_{i=1}^{n} u_i \right\|
\end{equation}

For vectors with varying magnitudes, such as Q/K vectors in attention heads, we generalize this to:
\begin{equation}
    R = \frac{\|\mathbb{E}[q]\|}{\mathbb{E}[\|q\|]}
\end{equation}
where expectations are taken over token positions. This ratio has intuitive bounds:
\begin{itemize}
    \item $R = 1$: All vectors point in exactly the same direction (perfect concentration)
    \item $R = 0$: Vectors are uniformly distributed in all directions (no concentration)
\end{itemize}

In practice, we compute $R_f$ for each frequency band $f$ separately:
\begin{equation}
    R_f = \frac{\|\mathbb{E}[q_f]\|}{\mathbb{E}[\|q_f\|]}
\end{equation}
High $R_f$ indicates that the trigonometric series approximation is accurate for band $f$, justifying the use of $S_{\text{trig}}$. The weighting factor $(1 - R_f)$ in $S_{\text{norm}}$ ensures that norm-based scoring contributes more when concentration is lower.

\subsection{Reconstruction Correlation}
\label{app:reconstruction-correlation}

To validate that Q/K concentration enables predictable distance preferences, we measure how well the trigonometric series reconstructs actual attention patterns. We define the \textbf{Reconstruction Correlation} $\bar{r}$ as follows.

For a given attention head, let $\hat{s}(\Delta)$ denote the predicted attention logit at distance $\Delta$, computed from Q/K centers via the trigonometric series (Equation~\ref{eq:reconstruction} in the main text). For each query $i$, let $\mathbf{a}_i = (a_{i,1}, a_{i,2}, \ldots)$ be its actual attention logits at distances $\Delta_1, \Delta_2, \ldots$, and let $\hat{\mathbf{s}} = (\hat{s}(\Delta_1), \hat{s}(\Delta_2), \ldots)$ be the corresponding predictions.

The per-query correlation is the Pearson correlation coefficient:
\begin{equation}
    r_i = \rho(\mathbf{a}_i, \hat{\mathbf{s}}) = \frac{\text{Cov}(\mathbf{a}_i, \hat{\mathbf{s}})}{\sigma_{\mathbf{a}_i} \sigma_{\hat{\mathbf{s}}}}
\end{equation}

The reconstruction correlation $\bar{r}$ is the average over all queries:
\begin{equation}
    \bar{r} = \frac{1}{N} \sum_{i=1}^{N} r_i
\end{equation}

\textbf{Distance Sampling.} To ensure balanced coverage across distance scales, we sample at logarithmically-spaced distances $\Delta \in \{1, 2, 4, 8, 16, \ldots\}$. This prevents nearby distances (which are numerous) from dominating the correlation and ensures that long-range attention patterns are adequately represented.

\subsection{Dominant Frequency Band Selection}
\label{app:dominant-bands}

Not all frequency bands contribute equally to the attention logit. We define \textit{dominant bands} as those contributing the most to the expected attention score. For each head, we compute the expected contribution of band $f$ as:
\begin{equation}
    C_f = \mathbb{E}[\|q_f\|] \cdot \mathbb{E}[\|k_f\|]
\end{equation}
where expectations are taken over a calibration dataset.

We rank bands by $C_f$ and select the Top-$K$ bands (typically $K=2$) for visualization. These dominant bands account for the majority of the attention logit magnitude. Figures in the main text visualize Q/K distributions in the 2D complex planes corresponding to these dominant bands.

\begin{figure}[t]
  \begin{center}
    \includegraphics[width=0.8\textwidth]{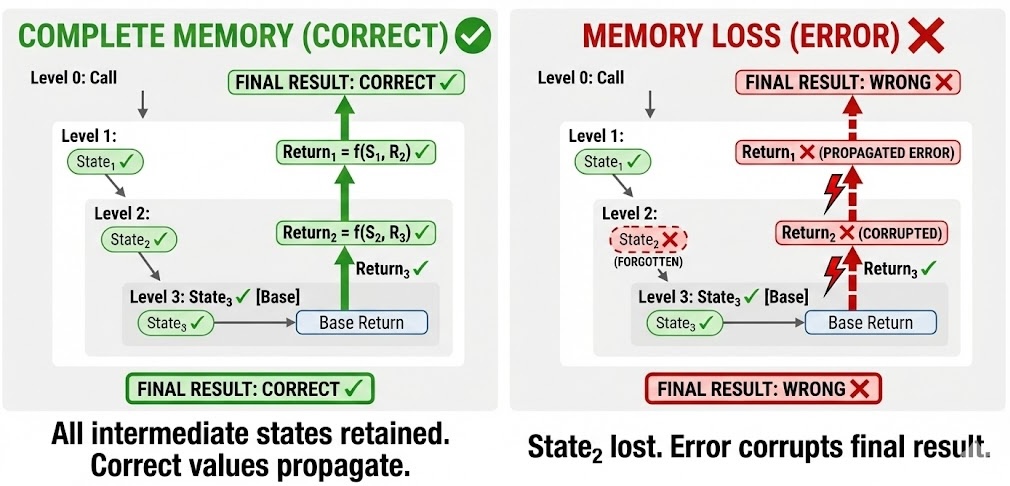}
    \caption{Evaluating memory via recursive simulation. \textbf{Left}: With complete memory, all intermediate states are retained and correct values propagate upward. \textbf{Right}: When an intermediate state is lost (State$_2$), the error propagates through all subsequent return values, corrupting the final result.}
    \label{fig:dfs_illustration}
  \end{center}
\end{figure}
\section{Recursive State Query Benchmark Details}
\label{app:dfs-benchmark}

This appendix provides details of the Recursive State Query benchmark used in \S\ref{subsubsec:memory}.

\subsection{Task Definition}

Given an undirected graph $G = (V, E)$ and a target step count $k$, the model simulates $k$ steps of depth-first search starting from a designated node and reports:
\begin{itemize}
    \item \textbf{Current node}: The node where the traversal currently resides
    \item \textbf{Stack state}: The complete path from the start node to the current node (in order)
    \item \textbf{Visited nodes}: All nodes that have been visited during the traversal
\end{itemize}

\subsection{Why DFS for Memory Evaluation}

DFS is well-suited for evaluating memory retention because:
\begin{itemize}
    \item \textbf{History dependency}: The stack state depends on the complete traversal history---any intermediate information loss causes errors
    \item \textbf{Uniform distribution}: Information is uniformly distributed across the sequence rather than concentrated at the beginning or end
    \item \textbf{Controllable difficulty}: Task difficulty scales directly with step count
    \item \textbf{Deterministic}: The algorithm is deterministic (neighbors selected in ascending order), providing unique ground truth
\end{itemize}

\subsection{Evaluation Metric}

We use \textit{stack exact match} as the primary metric, which requires the complete path to be correct in order. This is the strictest metric and most sensitive to information loss caused by KV cache pruning.

\subsection{Why Recursion Tests Memory}

Figure~\ref{fig:dfs_illustration} illustrates why recursive simulation effectively tests memory retention. Recursive algorithms require the model to first descend through nested calls, then backtrack to produce results. During backtracking, the model must recall intermediate states from earlier in the sequence. If any state is forgotten, the error propagates through all subsequent return values, corrupting the final result.

\begin{figure*}[t]
  \vskip 0.2in
  \begin{center}
    \includegraphics[width=0.95\textwidth]{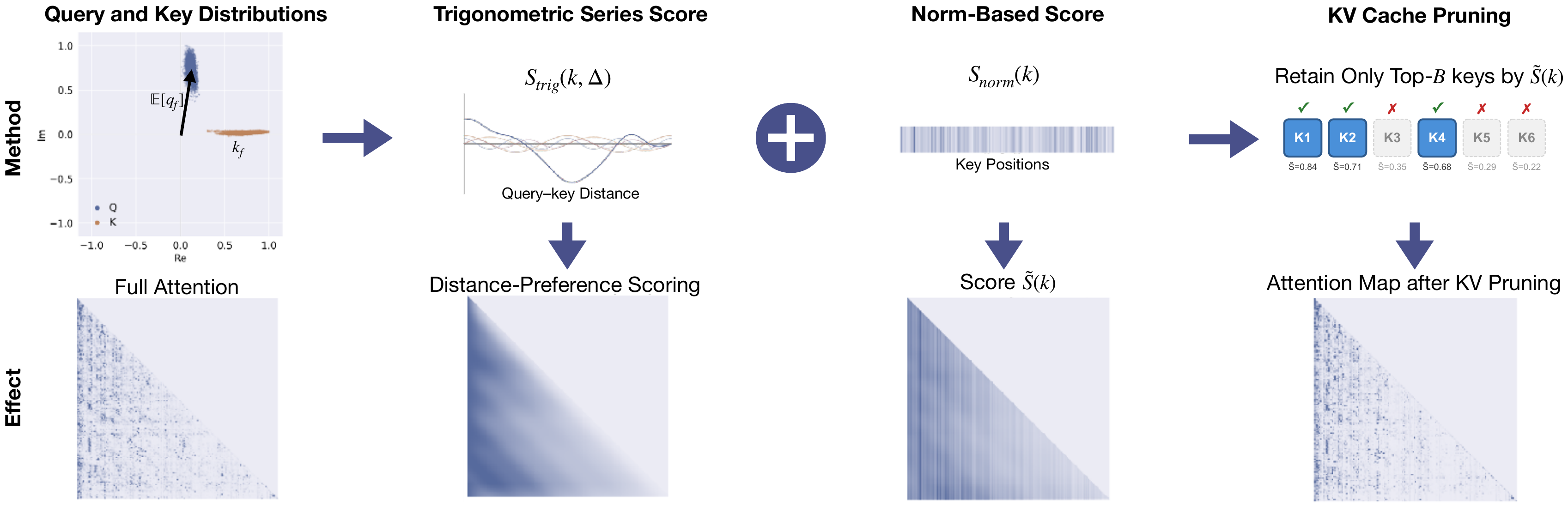}
    \caption{Method visualization with real attention maps, corresponding to the schematic in Figure~\ref{fig:method-overview}. \textbf{Top row}: The four stages of TriAttention. (1) We compute the Q/K centers $\mathbb{E}[q]$, $\mathbb{E}[k]$ from pre-RoPE distributions. (2) Using the trigonometric series, we compute $S_{\text{trig}}$ which scores keys based on distance preference. (3) We add the norm-based score $S_{\text{norm}}$, weighted by concentration, to obtain the final score $\bar{S}(k)$. (4) We retain top-scoring keys and evict the rest. \textbf{Bottom row}: Real attention maps illustrating the scenario described in Figure~\ref{fig:method-overview}. From left to right: original attention pattern showing distance preference; $S_{\text{trig}}$ visualization capturing the diagonal structure; combined score $\bar{S}$ incorporating norm information; attention after KV cache pruning, preserving the essential pattern.}
    \label{fig:method-visualization}
  \end{center}
  \vskip -0.2in
\end{figure*}

\section{Method Visualization}
\label{app:method-visualization}

Figure~\ref{fig:method-visualization} provides a detailed visualization of the TriAttention pipeline with real attention maps, complementing the schematic overview in Figure~\ref{fig:method-overview}.

\section{Comparison with Additional Baselines}
\label{app:general-tasks}

We compare TriAttention with additional KV cache compression methods beyond SnapKV and R-KV evaluated in the main text.

Table~\ref{tab:lazyeviction} compares with LazyEviction~\cite{zhang2025lazyevictionlaggedkveviction} on AIME24 using DeepSeek-R1-Distill-Qwen-7B at multiple KV budgets, alongside H2O~\cite{zhang2023h2o}, TOVA~\cite{oren2024tova}, and RaaS~\cite{hu2025raas} (results cited from LazyEviction). TriAttention outperforms all methods at every budget, and matches Full Attention at 30\% KV budget (46.7\%).

\begin{table}[h]
    \caption{Comparison on AIME24 (DeepSeek-R1-Distill-Qwen-7B) at varying KV budgets. $^*$Results cited from LazyEviction.}
    \centering
    \begin{tabular}{l|ccc}
        \toprule
        \textbf{Method} & \textbf{10\%} & \textbf{20\%} & \textbf{30\%} \\
        \midrule
        FullKV & \multicolumn{3}{c}{46.7} \\
        \midrule
        H2O$^*$ & -- & -- & 33.3 \\
        TOVA$^*$ & -- & -- & 36.7 \\
        RaaS$^*$ & -- & -- & 36.7 \\
        R-KV$^*$ & -- & -- & 43.3 \\
        LazyEviction & 33.3 & 40.0 & 43.3 \\
        \rowcolor{lm_purple} TriAttention & \textbf{40.0} & \textbf{43.3} & \textbf{46.7} \\
        \bottomrule
    \end{tabular}
    \label{tab:lazyeviction}
\end{table}

\section{Evaluation on Diverse Benchmarks}
\label{app:diverse-benchmarks}

To evaluate generalization beyond mathematical reasoning, we test on LongBench~\cite{bai2024longbench} (16 subtasks spanning QA, summarization, dialogue, retrieval, and code; Qwen3-8B, 50\% KV budget) and RULER~\cite{hsieh2024ruler} (retrieval tasks, 4K context). We compare against StreamingLLM~\cite{xiao2024efficient}, PyramidKV~\cite{cai2024pyramidkv}, KnormPress~\cite{devoto2025expectedattention}, Ada-KV+SnapKV~\cite{feng2025adakv}, and H2O~\cite{zhang2023h2o}.

Table~\ref{tab:longbench_full} presents the full LongBench results. TriAttention achieves the highest average (48.1) across 16 subtasks, winning 11 out of 16, and surpasses Ada-KV+SnapKV by +2.5 (48.1 vs.\ 45.6). On RULER (Table~\ref{tab:ruler}), TriAttention outperforms all baselines with 66.1, a +10.5 gap over SnapKV. We also compare with H2O, which requires $O(n^2)$ memory and cannot use FlashAttention; on the 12 LongBench subtasks where H2O fits in 48GB GPU memory (Table~\ref{tab:longbench_h2o}), TriAttention wins 10 out of 12.

\begin{table*}[t]
    \caption{LongBench results (Qwen3-8B, 50\% KV budget). 16 subtasks spanning QA, summarization, few-shot classification, retrieval, counting, and code. \textbf{Bold} = best among compression methods.}
    \centering
    \resizebox{\textwidth}{!}{
    \begin{tabular}{l|cccccc|ccc|ccc|c|c|cc|c}
        \toprule
        & \multicolumn{6}{c|}{\textbf{QA}} & \multicolumn{3}{c|}{\textbf{Summarization}} & \multicolumn{3}{c|}{\textbf{Few-shot}} & \textbf{Retr.} & \textbf{Count} & \multicolumn{2}{c|}{\textbf{Code}} & \\
        \cmidrule(lr){2-7} \cmidrule(lr){8-10} \cmidrule(lr){11-13} \cmidrule(lr){14-14} \cmidrule(lr){15-15} \cmidrule(lr){16-17}
        \textbf{Method} & NarrQA & Qasp & MFQA & HpQA & 2Wik & Musi & GovR & QMSu & MNew & TREC & TriQA & SSum & PaRe & PaCn & LCC & ReBe & \textbf{Avg} \\
        \midrule
        Full Attn. & 28.8 & 43.8 & 55.3 & 62.8 & 48.9 & 35.5 & 33.5 & 24.7 & 24.7 & 40.5 & 90.5 & 40.3 & 91.8 & 9.0 & 64.9 & 60.0 & 47.2 \\
        \midrule
        SnapKV & 26.9 & 37.0 & 45.5 & 59.2 & \textbf{46.3} & 33.1 & 32.2 & 23.0 & 23.4 & 40.5 & 89.9 & \textbf{41.0} & \textbf{91.1} & 7.5 & \textbf{66.4} & 59.9 & 45.2 \\
        PyramidKV & 25.9 & 30.4 & 39.1 & 52.2 & 39.9 & 29.7 & 29.9 & 22.2 & 21.9 & 33.5 & 90.0 & 40.7 & 93.4 & 8.0 & 65.7 & 60.2 & 42.7 \\
        StreamingLLM & 24.1 & 30.5 & 31.2 & 46.5 & 41.6 & 21.6 & 30.8 & 21.8 & 23.9 & 43.0 & 85.4 & 38.2 & 55.2 & \textbf{10.0} & 64.7 & 61.2 & 39.4 \\
        KnormPress & 17.6 & 24.4 & 40.3 & 29.2 & 26.4 & 14.9 & 28.8 & 21.6 & 20.9 & 50.0 & 81.6 & 41.2 & 79.1 & 7.1 & 31.7 & 47.5 & 35.1 \\
        Ada-KV+SnapKV & 27.0 & 36.8 & 45.0 & 59.6 & \textbf{47.5} & 34.1 & 31.8 & 23.0 & 23.6 & 46.0 & 90.1 & \textbf{40.9} & 90.8 & \textbf{8.0} & 65.6 & 59.6 & 45.6 \\
        \rowcolor{lm_purple} TriAttention & \textbf{28.1} & \textbf{43.0} & \textbf{51.4} & \textbf{60.2} & 44.9 & \textbf{36.9} & \textbf{32.9} & \textbf{23.8} & \textbf{24.3} & \textbf{69.0} & \textbf{90.3} & 39.9 & \textbf{91.0} & 7.0 & \textbf{65.0} & \textbf{61.3} & \textbf{48.1} \\
        \bottomrule
    \end{tabular}
    }
    \label{tab:longbench_full}
\end{table*}

\begin{table}[h]
    \caption{RULER retrieval results (Qwen3-8B, 50\% KV, 4K context). \textbf{Bold} = best among compression methods.}
    \centering
    \begin{tabular}{l|c}
        \toprule
        \textbf{Method} & \textbf{RULER Avg} \\
        \midrule
        SnapKV & 55.6 \\
        PyramidKV & 40.7 \\
        StreamingLLM & 61.1 \\
        \rowcolor{lm_purple} TriAttention & \textbf{66.1} \\
        \bottomrule
    \end{tabular}
    \label{tab:ruler}
\end{table}

\begin{table*}[t]
    \caption{Comparison with H2O on LongBench subtasks where H2O fits in 48GB GPU memory (Qwen3-8B, 50\% KV). H2O requires materializing the full $O(n^2)$ attention matrix and cannot use FlashAttention. \textbf{Bold} = best.}
    \centering
    \resizebox{\textwidth}{!}{
    \begin{tabular}{l|cccccccccccc|c}
        \toprule
        \textbf{Method} & Qasp & HpQA & 2Wik & Musi & GovR & QMSu & MNew & TREC & TriQA & SSum & NarrQA & MFQA & \textbf{Avg} \\
        \midrule
        H2O & 39.2 & 50.7 & 43.9 & 30.7 & \textbf{32.9} & 23.4 & \textbf{24.4} & 56.5 & 89.1 & 39.1 & 21.2 & 45.4 & 41.4 \\
        \rowcolor{lm_purple} TriAttention & \textbf{43.0} & \textbf{60.2} & \textbf{44.9} & \textbf{36.9} & \textbf{32.9} & \textbf{23.8} & 24.3 & \textbf{69.0} & \textbf{90.3} & \textbf{39.9} & \textbf{28.1} & \textbf{51.4} & \textbf{45.4} \\
        \bottomrule
    \end{tabular}
    }
    \label{tab:longbench_h2o}
\end{table*}

\section{Future Offset Design Ablation}
\label{app:future-offset}

Our scoring function evaluates key importance at multiple future offsets $\mathcal{D}$. We ablate two aspects: the range/number of offsets, and the spacing strategy (Table~\ref{tab:ablation_offset}).

\begin{table}[h]
    \caption{Future offset ablation on Qwen3-8B (AIME24). Top: effect of offset range. Bottom: spacing strategy comparison (17 offsets, range [1, 65536]).}
    \centering
    \begin{tabular}{cc|c}
        \toprule
        \textbf{Max Dist} & \textbf{\#Offsets} & \textbf{Acc} \\
        \midrule
        128 & 8 & 41.7 \\
        4096 & 13 & \textbf{48.8} \\
        8192 & 14 & 46.2 \\
        65536 & 17 & 45.8 \\
        \midrule
        \multicolumn{2}{l|}{Linear spacing} & 28.7 \\
        \multicolumn{2}{l|}{\cellcolor{lm_purple}Geometric spacing} & \cellcolor{lm_purple}\textbf{45.8} \\
        \bottomrule
    \end{tabular}
    \label{tab:ablation_offset}
\end{table}

Increasing the maximum distance from 128 to 4096 improves accuracy from 41.7\% to 48.8\% (+7.1\%), confirming that future offsets are beneficial. Geometric spacing ($\{1, 2, 4, \ldots\}$) dramatically outperforms linear spacing (45.8\% vs.\ 28.7\%, $-17.1\%$), as near-distance positions require denser sampling.

\section{Calibration Data Sensitivity}
\label{app:calib-sensitivity}

We test robustness to calibration data quantity and quality (Table~\ref{tab:ablation_calib_sensitivity}).

\begin{table}[h]
    \caption{Calibration data sensitivity on Qwen3-8B (AIME24). Top: effect of calibration data size. Bottom: effect of calibration data quality.}
    \centering
    \begin{tabular}{l|c}
        \toprule
        \textbf{Calibration} & \textbf{Acc} \\
        \midrule
        50k tokens & 45.4 \\
        200k tokens & 45.8 \\
        960k tokens & 45.8 \\
        \midrule
        HTML (low qual.) & 46.2 \\
        Code (mid qual.) & 43.3 \\
        Chat (high qual.) & 46.7 \\
        \bottomrule
    \end{tabular}
    \label{tab:ablation_calib_sensitivity}
\end{table}

Performance is stable across calibration sizes from 50k to 960k tokens (45.4--45.8\%). Similarly, calibration data quality shows no clear correlation with accuracy: using Google homepage HTML (low quality) achieves 46.2\%, comparable to ShareGPT chat data (46.7\%). This confirms that the Q/K statistics captured during calibration are model-intrinsic properties, robust to the choice of calibration data.

\section{MLA Architecture Validation}
\label{app:mla}

To test whether Q/K concentration generalizes beyond standard attention architectures, we evaluate on GLM-4.7-Flash, which uses Multi-head Latent Attention (MLA) with 940 heads. Table~\ref{tab:mla} compares reconstruction quality (Pearson $r$) and directional concentration (MRL) between GQA and MLA architectures.

\begin{table}[t]
    \caption{Cross-architecture comparison of reconstruction quality (Pearson $r$) and Q/K concentration (MRL) between GQA (Qwen3-8B) and MLA (GLM-4.7-Flash) architectures.}
    \centering
    \begin{subtable}[t]{0.48\textwidth}
        \centering
        \caption{Reconstruction quality (Pearson $r$)}
        \begin{tabular}{l|cc}
            \toprule
            \textbf{Threshold} & \textbf{Qwen3-8B (GQA)} & \textbf{GLM-4.7 (MLA)} \\
            \midrule
            $r > 0.90$ & 0.8\% & 1.7\% \\
            $r > 0.70$ & 13.0\% & 23.1\% \\
            $r > 0.50$ & 53.5\% & 51.6\% \\
            \bottomrule
        \end{tabular}
    \end{subtable}
    \hfill
    \begin{subtable}[t]{0.48\textwidth}
        \centering
        \caption{Q/K Concentration (MRL)}
        \begin{tabular}{l|cc}
            \toprule
            \textbf{Threshold} & \textbf{Qwen3-8B (GQA)} & \textbf{GLM-4.7 (MLA)} \\
            \midrule
            $R > 0.95$ & 84.7\% & 96.6\% \\
            $R > 0.90$ & 90.8\% & 99.8\% \\
            \bottomrule
        \end{tabular}
    \end{subtable}
    \label{tab:mla}
\end{table}

The MLA architecture shows comparable or stronger concentration and reconstruction quality compared to GQA. Notably, 96.6\% of MLA heads achieve $R > 0.95$ (vs.\ 84.7\% for GQA), indicating that the Q/K concentration phenomenon is architecture-general and potentially even more pronounced in MLA.

\section{Real-World Deployment: Agentic Task on a Single GPU}
\label{app:demo}

We demonstrate TriAttention in a practical deployment scenario using OpenClaw, a multi-turn agent. We serve Qwen3-32B with AWQ INT4 quantization on a single RTX 4090 (24GB), where model weights alone consume most of the GPU memory, leaving a very limited budget for KV cache. This makes the setup particularly challenging for OpenClaw, since its prompt in the first request has already exceeds 15k tokens, and each interaction round further expands the context as the agent reads and processes six markdown documents to produce a weekly report.

\begin{figure}[t]
    \centering
  \includegraphics[width=0.9\columnwidth]{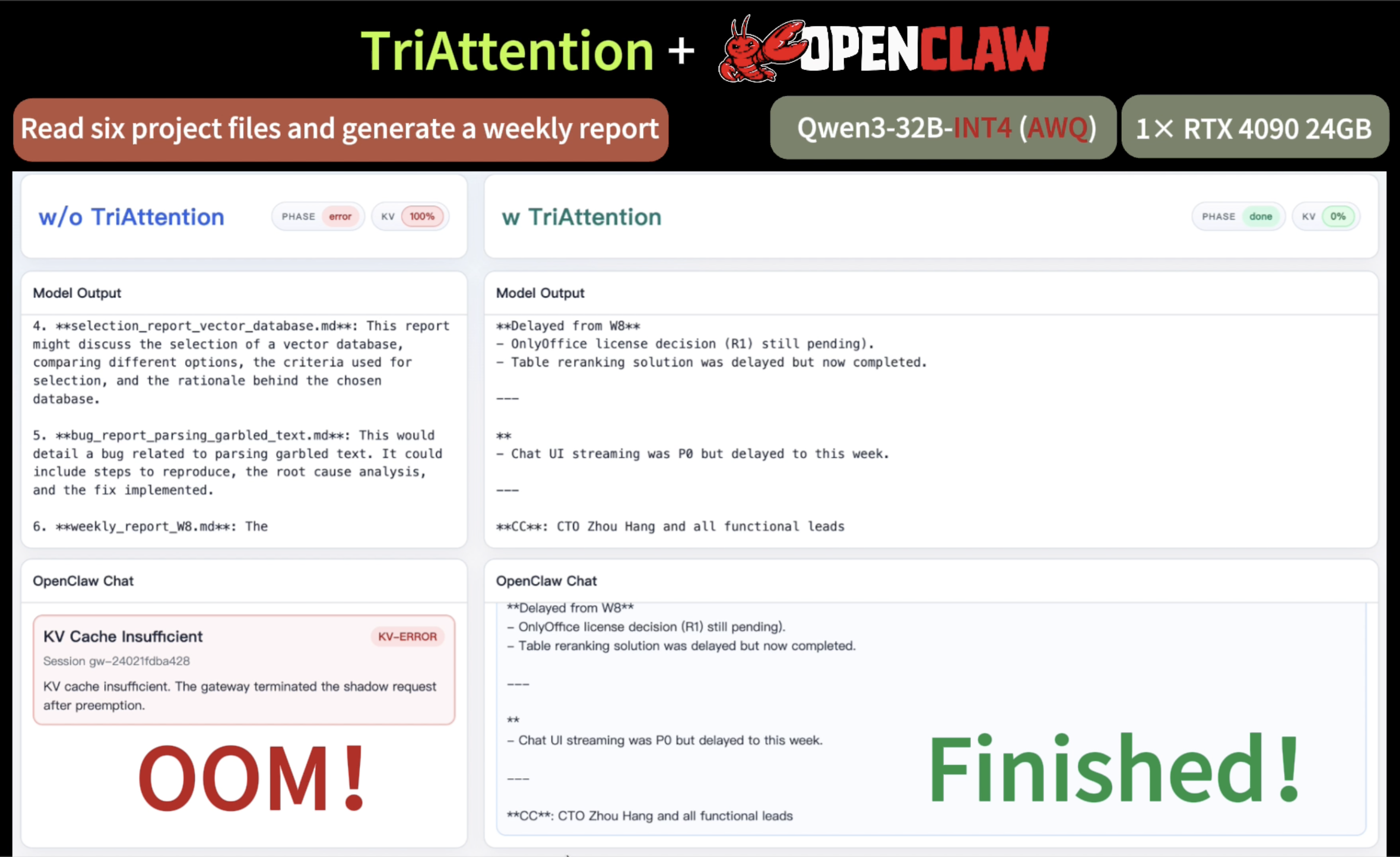}    
    \caption{OpenClaw demo on a single RTX 4090 with Qwen3-32B (INT4). Full attention runs out of memory during multi-turn interaction, while TriAttention completes the task within the GPU memory budget. Full video is available on our GitHub page: \url{https://github.com/WeianMao/triattention}.}
    \label{fig:demo}
\end{figure}

With full attention (baseline), the KV cache grows unboundedly during multi-turn interaction, causing an out-of-memory error before the agent can complete the task. With TriAttention, KV cache compression keeps memory usage within budget throughout the entire session, allowing the agent to successfully read all documents and generate the report. A screenshot of the completed session is shown in Figure~\ref{fig:demo}.

\end{document}